\documentclass[11pt]{article}

\usepackage[T1]{fontenc}
\usepackage[utf8]{inputenc}
\usepackage[a4paper,margin=1in]{geometry}
\usepackage{microtype} 

\usepackage{amsmath,amssymb,amsfonts}
\numberwithin{equation}{section}

\usepackage{graphicx}
\graphicspath{{figures/}}
\usepackage{booktabs}
\usepackage{caption}
\usepackage{subcaption}

\usepackage[numbers,sort&compress]{natbib}
\usepackage{url}
\usepackage[hidelinks]{hyperref}
\usepackage[capitalise,noabbrev]{cleveref}

\usepackage{authblk}

\title{A Foundation Model for Brain MRI with Dynamic Modality Integration}
\author[1]{Minh Sao Khue Luu}
\author[1]{Bair N. Tuchinov}
\affil[1]{The Artificial Intelligence Research Center of Novosibirsk State University, Novosibirsk 630090, Russia}
\affil[ ]{\texttt{khue.luu@g.nsu.ru}}
\date{}

\begin{document}
\maketitle

\begin{abstract}
We present a foundation model for brain MRI that can work with different combinations of imaging sequences. The model uses one encoder with learnable modality embeddings, conditional layer normalization, and a masked autoencoding objective that accounts for missing modalities. A variance–covariance regularizer is applied to stabilize feature learning and improve representation diversity. This design removes the need for separate models for each modality and allows the network to adapt when some sequences are missing or unseen. It is trained on about 60{,}000 multi-center MRIs using self-supervised reconstruction and modality imputation to learn flexible representations. A learnable modality embedding 
guides feature extraction so the encoder can adjust to different inputs. We describe our planned evaluation on brain tumor and multiple sclerosis segmentation, as well as lesion classification, under various modality settings. Preliminary 
results show that the method works feasibly, and further experiments are planned to study its performance in more detail. All code and pretrained models are available at \url{https://github.com/BrainFM/brainfm}
\end{abstract}


\section{Introduction}
Foundation models (FMs) have transformed natural language processing, computer vision, and multimodal learning, yet their application to medical imaging remains challenging. Brain MRI relies on multiple sequences (e.g., T1, T2, FLAIR), but acquisition protocols vary widely across sites. This variation leads to inconsistent modality availability and limits the generalization of fixed-modality architectures \cite{wang_mmformer_2022}.

Masked autoencoders and foundation models have demonstrated strong capabilities for learning general-purpose visual representations \cite{he_masked_2021}. In medical imaging, self-supervised pretraining approaches such as Models Genesis 
\cite{zhou_models_2020} have shown that large-scale unsupervised learning can reduce annotation costs and improve cross-dataset generalization. Recent work extends these ideas to multimodal MRI: AMAES \cite{munk_amaes_2024} pretrains on 
44k MRIs using masked autoencoding but supports only single-modality inputs; M4oE \cite{linguraru_m4oe_2024} and MoME \cite{linguraru_foundation_2024} use modality-specific experts with gating to enable dynamic fusion but require new experts for additional sequences. mmFormer \cite{wang_mmformer_2022} addresses missing MRI modalities for brain tumor segmentation using modality-specific and cross-modal Transformers with regularization for robust predictions.

Existing approaches such as AMAES are limited to single-modality inputs, while MoME and M4oE rely on separate expert networks for each sequence. Our model uses a single backbone combined with learnable modality embeddings, allowing it to handle new sequences without adding new experts. Compared with mmFormer, which focuses on task-specific robustness, we pretrain a unified encoder with modality imputation for broader downstream applications. The proposed foundation model 
for brain MRI (a) conditions feature extraction on learnable text-derived modality embeddings, (b) adapts shared encoder parameters through conditional layer normalization, and (c) integrates self-supervised modality imputation within a masked reconstruction framework.The design removes the need for modality-specific branches and maintains robustness when sequences are missing or previously unseen. A variance–covariance regularizer further stabilizes learning across variable modality sets, enabling consistent performance across diverse MRI configurations.

\section{Method}
\subsection{Dataset Preparation and Preprocessing}

\paragraph{Pretraining Dataset.}
We use the FOMO25 dataset \cite{munk_large-scale_2025} for pretraining. It includes 11,187 subjects, 13,900 sessions, and 60,529 MRI scans in total. Each subject may undergo one or more imaging sessions, and each session contains several MRIs saved in the NIfTI format. In this study, we treat a \emph{case} at the session level: for example, if a subject has two sessions, these are considered as two separate cases. 

\paragraph{Training Dictionary.}
To manage the data efficiently, we organize all subject sessions in a nested dictionary (subjects with multiple sessions have distinct keys). The outer dictionary uses \texttt{case\_id} as the key, and its value is another dictionary that maps each modality to the corresponding NIfTI image, i.e., \texttt{\{case\_id : \{modality : path\}\}}. This structure allows flexible handling of cases with different numbers or types of modalities. This nested structure allows flexible access to heterogeneous modality combinations across sessions.

\paragraph{Preprocessing Pipeline.}
During data loading, all modalities of a given session are read from disk and cropped or padded to a consistent spatial size of $(128,128,128)$ (voxels). A sequence of 3D data augmentations is then applied using the \texttt{MONAI} library. Table~\ref{tab:preprocess} summarizes all preprocessing operations and their hyperparameters. To improve efficiency, text embeddings of modality names are cached using a modality encoder cache, ensuring that identical modality tokens share the same embedding across samples.

\begin{table}[h]
\centering
\caption{Preprocessing and augmentation operations applied during dataset loading.}
\label{tab:preprocess}
\begin{tabular}{p{4cm}p{9cm}}
\toprule
\textbf{Operation} & \textbf{Description and Hyperparameters} \\
\midrule
\texttt{DivisiblePad} & Pads volumes to make each spatial dimension divisible by the patch size $(16,16,16)$. \\
\texttt{RandBiasField} & Random bias field distortion; probability $p=0.3$, coefficient range $(0.3, 0.6)$. \\
\texttt{RandGaussianNoise} & Adds Gaussian noise; probability $p=0.3$, mean $0.0$, standard deviation $0.05$. \\
\texttt{RandAdjustContrast} & Random contrast adjustment; probability $p=0.3$, gamma range $(0.7, 1.5)$. \\
\texttt{RandFlip} & Random flipping along all spatial axes; probability $p=0.5$. \\
\texttt{RandAffine} & Random rotation within $\pm15^\circ$ about each axis ($\pm\pi/12$ radians); 
padding mode: \texttt{border}. \\
\texttt{NormalizeIntensity} & Normalizes nonzero voxels per channel to zero mean and unit variance. \\
\texttt{Sanitize Voxels} & Replaces non-finite values (NaN, Inf) with zeros and clamps range to $[-4,4]$. \\
\bottomrule
\end{tabular}
\end{table}

\subsection{Architecture}

BrainFM is a modality-conditioned masked autoencoder designed for large-scale 3D brain MRI pretraining. Its design unifies four key ideas: (1) learnable text-derived modality embeddings that generalize to unseen sequences, (2) conditional layer normalization (CLN) for adaptive feature modulation across modalities, (3) padding-aware masking for handling variable input completeness, and (4) variance--covariance regularization
for stable and diverse representation learning.

\begin{figure}[h]
    \centering
    \includegraphics[width=0.8\linewidth]{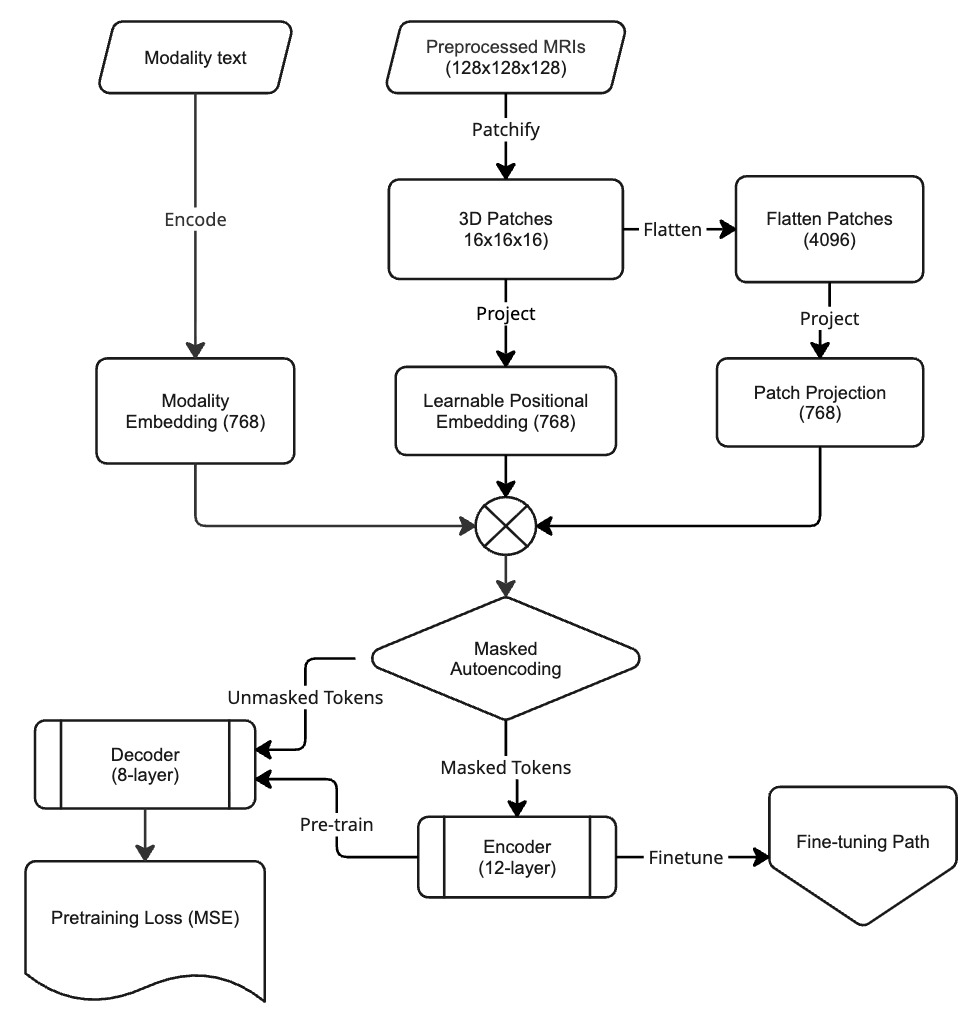}
    \caption{
    Overview of the proposed modality-conditioned masked autoencoder for 3D brain MRI. Visible (unmasked) patches are encoded, and masked patches are reconstructed in the decoder. The workflow includes patch extraction,
    modality conditioning, masking, and reconstruction.
    }
    \label{fig:fig_architecture}
\end{figure}

\paragraph{Overview.}
Each MRI volume is divided into non-overlapping 3D patches that are flattened and linearly projected into tokens. Each token is enriched with a modality embedding (obtained from a pretrained text encoder) and a learnable 3D positional embedding. The modality embedding is injected at each encoder layer through CLN, allowing modality-specific feature scaling and shifting during representation learning. This conditioning allows the encoder to adaptively modulate its normalization statistics depending on the modality. A padding-aware masking strategy (randomly hides a fixed proportion $\rho$ of valid patches while ensuring that padded or empty regions are never selected for masking) is then applied. The visible tokens are processed by a Transformer encoder with CLN, and the masked tokens are reconstructed by a lightweight decoder using modality-aware mask tokens and positional re-injection, as shown in Figure~\ref{fig:fig_architecture}.

\paragraph{Masked Reconstruction Loss.}
The model is first trained to reconstruct the hidden (masked) patches from the visible ones. The objective is a mean squared error (MSE) loss computed only on masked and valid voxel elements:

\begin{equation}
\mathcal{L}_{\mathrm{MAE}}
= \frac{1}{N}
\sum_{s}\sum_{i \in \mathcal{H}_s}\sum_{p}
\big(\widehat{P}_{s,i}^{(p)} - P_{s,i}^{(p)}\big)^2,
\end{equation}

where $P_{s,i}^{(p)}$ and $\widehat{P}_{s,i}^{(p)}$ denote original and reconstructed voxel intensities, $\mathcal{H}_s$ are masked patch indices, and $N$ is the number of valid voxel elements. This reconstruction task encourages the model to learn modality-aware and spatially coherent features across heterogeneous MRI inputs.

\paragraph{Variance--Covariance Regularization.}
To further stabilize pretraining and avoid feature collapse, BrainFM-MRI applies a VICReg-inspired regularization ~\cite{bardes_vicreg_2021}. Each
sample produces a mean-pooled feature vector
$\mathbf{z}_b \in \mathbb{R}^{D}$, stacked across the batch as
$\mathbf{z} \in \mathbb{R}^{B\times D}$. Two auxiliary losses are added:
the \emph{variance term}, which enforces unit variance per feature dimension, and the \emph{covariance term}, which penalizes correlations between features:

\begin{equation}
\mathcal{L}_{\mathrm{var}} =
\frac{1}{D}\sum_{j}
\mathrm{ReLU}\!\left(1-\sqrt{\mathrm{Var}(z_{\cdot j})+\epsilon}\right)
\end{equation}

\begin{equation}
\mathcal{L}_{\mathrm{cov}} =
\frac{1}{D(D-1)}\sum_{i\neq j}
\mathrm{Cov}(z_{\cdot i},z_{\cdot j})^2.
\end{equation}

\paragraph{Total Objective.}
The final pretraining objective combines reconstruction and
regularization terms:

\begin{equation}
\mathcal{L}_{\mathrm{total}} =
\mathcal{L}_{\mathrm{MAE}} +
\lambda_{\mathrm{var}}\mathcal{L}_{\mathrm{var}} +
\lambda_{\mathrm{cov}}\mathcal{L}_{\mathrm{cov}},
\end{equation}

where $\lambda_{\mathrm{var}}=0.1$ and
$\lambda_{\mathrm{cov}}=0.005$ are linearly increased during the first five epochs (warm-up). 

\paragraph{Transfer to Downstream Tasks.}
After pretraining, the decoder is discarded, and the encoder is adapted to downstream segmentation or classification tasks. A lightweight head is attached—mean-pooled for classification or upsampled for segmentation— and trained with a standard supervised loss $\mathcal{L}_{\mathrm{task}}$.

\section{Experimental Plan}
\label{sec:experiments}
\paragraph{Models.}
We compare BrainFM against AMAES, MoME, M4oE, and mmFormer. All models use identical preprocessing, augmentations, and compute budgets for fairness.

\paragraph{Datasets.}
We evaluate on two segmentation datasets (e.g., BraTS-MET, MSLesSeg) and two classification datasets (e.g., lesion presence/severity or tumor subtype) plus one cross-domain dataset to probe shift. Each dataset is split subject-wise into train/val/test (70/10/20) with three fixed, non-overlapping folds; no subject leakage. All images are resampled to a common voxel spacing, intensity normalized (z-score), and optionally bias-field corrected. Spatial (flip, affine) and photometric (noise, contrast) augmentations are matched across models. Patch size, stride, and masking (for MAE-style methods) are harmonized wherever applicable.

\paragraph{Training Protocol.}
All models use AdamW, cosine decay with linear warm-up, the same batch size, epochs, and weight decay. Early stopping is based on validation Dice (segmentation) or AUROC (classification). Each run repeats with three seeds; we report mean~$\pm$~std. Hyperparameters are tuned on the
first fold’s validation split, then fixed for all other folds.

\paragraph{Evaluation Tasks.}
Segmentation: lesion/tumor segmentation with robustness to missing/unseen modalities and cross-domain generalization.
Classification: subject-level or scan-level labels (e.g., lesion presence, tumor subtype, disease vs.\ healthy), with:
(i) frozen-encoder linear probe,
(ii) shallow head finetuning, and
(iii) full-model finetuning (where permitted).

\paragraph{Modality-Availability Protocol.}
All models are evaluated under the configurations in
Table~\ref{tab:modality-configs}. ``Unseen'' rows are held out from training. For BrainFM-MRI, we additionally perform zero-shot tests on unseen modality sets to assess compositional generalization.

\begin{table}[t]
\centering
\small
\caption{Example modality configurations used for evaluation.
X indicates that the modality is available, and -- indicates it is missing.}
\label{tab:modality-configs}
\begin{tabular}{@{}lcccc@{}}
\toprule
Configuration            & T1 & T1c & T2 & FLAIR \\
\midrule
Complete                 & X  & X   & X  & X     \\
Dropped (T1c)            & X  & --  & X  & X     \\
Dropped (T2)             & X  & X   & -- & X     \\
Dropped (FLAIR)          & X  & X   & X  & --    \\
Unseen (T1+FLAIR only)   & X  & --  & -- & X     \\
Unseen (T2 only)         & -- & --  & X  & --    \\
\bottomrule
\end{tabular}
\end{table}

\section{Preliminary Results}
This small-scale experiment is an initial step toward the full benchmark described in Section~\ref{sec:experiments}. In this experiment, we pretrain the model using masked autoencoding with modality-conditioned reconstruction. For transfer, we consider two tasks: brain tumor segmentation and multiple sclerosis lesion analysis. Robustness is assessed using a Modality-Availability Matrix that includes full, missing, and unseen sequence configurations. Baselines include nnU-Net, SegResNet, and ablations without modality embeddings or imputation. Metrics follow standard practice (Dice, Hausdorff distance, sensitivity, specificity). Experiments are ongoing, and complete results will be reported in the final version.

At this stage, we verify the encoder’s feasibility on a small 10-case subset of the MSLesSeg dataset. Training was limited to two epochs using only the T1 and FLAIR modalities. Table~\ref{tab:prelim} shows early Dice and Hausdorff 
distance results, comparing the modality-conditioned MAE (BrainFM) with nnU-Net. These results illustrate early evidence that the modality-conditioned encoder learns stable representations even under limited supervision.

\begin{table}[h]
\centering
\small
\caption{Preliminary results on 10 MSLesSeg cases.}
\begin{tabular}{lcc}
\toprule
\textbf{Model} & \textbf{Dice} $\uparrow$ & \textbf{HD95} $\downarrow$ \\
\midrule
nnU-Net (T1+FLAIR) & 0.39 & 12.8 \\
BrainFM (ours)     & 0.45 & 11.2 \\
\bottomrule
\end{tabular}
\label{tab:prelim}
\end{table}

\section{Conclusion and Discussion}
We presented BrainFM-MRI, a modality-conditioned foundation model for brain MRI that unifies masked autoencoding and modality embeddings within a single encoder. Unlike prior expert-based architectures, BrainFM-MRI adapts dynamically to any set of MRI sequences, including missing or unseen ones. Table~\ref{tab:related_models} compares our approach with representative foundation models for brain MRI. Preliminary results indicate the feasibility of the approach, though experiments are still ongoing. Future work will
extend training to the full dataset and include comprehensive evaluations across multiple tasks.

\begin{table}[t]
\centering
\caption{Comparison of related foundation models.}
\label{tab:related_models}
\begin{tabular}{@{}lccccl@{}}
\toprule
\textbf{Model} & \textbf{Input} & \textbf{Handling} & \textbf{Scalability} \\ 
\midrule
AMAES   & Single & Masked AE & Limited \\
MoME    & Multi  & Experts + Gating & Needs new experts \\
M4oE    & Multi  & Mixture of Experts & Scales poorly \\
mmFormer & Multi & Cross-modal Transformer & Robust \\
Ours    & Arbitrary & Embeddings + CLN & Scalable \\
\bottomrule
\end{tabular}
\end{table}

\paragraph{Acknowledgements.}
This work was supported by a grant for research centers, provided by the Ministry of Economic Development of the Russian Federation in accordance with the subsidy agreement with the Novosibirsk State University dated April 17, 2025 No. 139-15-2025-006: IGK 000000C313925P3S0002

\bibliographystyle{unsrtnat}
\bibliography{references}

\end{document}